\pdfoutput=1

\documentclass[11pt]{article}

\usepackage{ACL2023}

\usepackage{times}
\usepackage{latexsym}
\usepackage[T1]{fontenc}
\usepackage[utf8]{inputenc}
\usepackage{microtype}
\usepackage{enumitem} 

\usepackage{inconsolata}
\usepackage{multirow}
\usepackage{multicol}
\usepackage{graphicx}
\usepackage{subcaption}
\usepackage{todonotes}
\usepackage{booktabs}
\usepackage{amsmath}
\usepackage{tipa}

\usepackage{graphicx}
\usepackage{subcaption}
\usepackage{xcolor}

\title{ 
PaperVoyager : Building  Interactive Web with Visual Language Models 
 
}

\author{
Dasen Dai$^{1*}$,
Biao Wu$^{2*}$,
Meng Fang$^{3}$,  
Wenhao Wang$^{1\dagger}$ \\
$^{1}$Vast Intelligence Lab, $^{2}$UTS, $^{3}$University of Liverpool \\
$^{*}$Equal contribution, $^{\dagger}$Corresponding author
}

\begin{document}
\maketitle
\begin{abstract}

Recent advances in visual language models have enabled autonomous agents for complex reasoning, tool use, and document understanding. However, existing document agents mainly transform papers into static artifacts such as summaries, webpages, or slides, which are insufficient for technical papers involving dynamic mechanisms and state transitions. In this work, we propose a Paper-to-Interactive-System Agent that converts research papers into executable interactive web systems. Given a PDF paper, the agent performs end-to-end processing without human intervention, including paper understanding, system modeling, and interactive webpage synthesis, enabling users to manipulate inputs and observe dynamic behaviors. To evaluate this task, we introduce a benchmark of 19 research papers paired with expert-built interactive systems as ground truth. We further propose PaperVoyager, a structured generation framework that explicitly models mechanisms and interaction logic during synthesis. Experiments show that PaperVoyager significantly improves the quality of generated interactive systems, offering a new paradigm for interactive scientific paper understanding.


\end{abstract}

\section{Introduction}

Recent advances in large language models (LLMs), such as Claude Code~\citep{anthropic_claude_code_2025} and GPT-5~\citep{singh2025openai}, have significantly improved AI capabilities in reasoning, tool use, and autonomous task execution~\citep{qin2023toolllm, zheng2024gpt,yan2023gpt4v,yao2022react,wu2026vision}. Building on these capabilities, recent work has explored document understanding systems that transform research papers into more accessible formats, such as structured summaries, Markdown documents, webpages, and slide decks~\citep{lu2025webgen,jiangwebgen,wang2025webgen,wang2025infinity,chen2025paper2web,shi2025presentagent}.

However, most existing approaches focus on static content transformation, simply converting documents into other formats while keeping the reading experience passive~\citep{chen2025paper2web,wang2025webgen,lu2025webgen}. In contrast, many technical papers describe mechanisms involving states, dynamics, and interactions, such as algorithm procedures and system pipelines, which are difficult to understand through static descriptions alone. Current document agents and webpage generators rarely provide interactive environments where users can manipulate inputs and observe system behaviors, limiting their ability to support deeper understanding of technical content~\citep{yao2022webshop,deng2023mind2web,nakano2021webgpt,zhou2023webarena,gur2023real}.

\begin{figure}[t!]
  \centering
  \includegraphics[width=\linewidth]{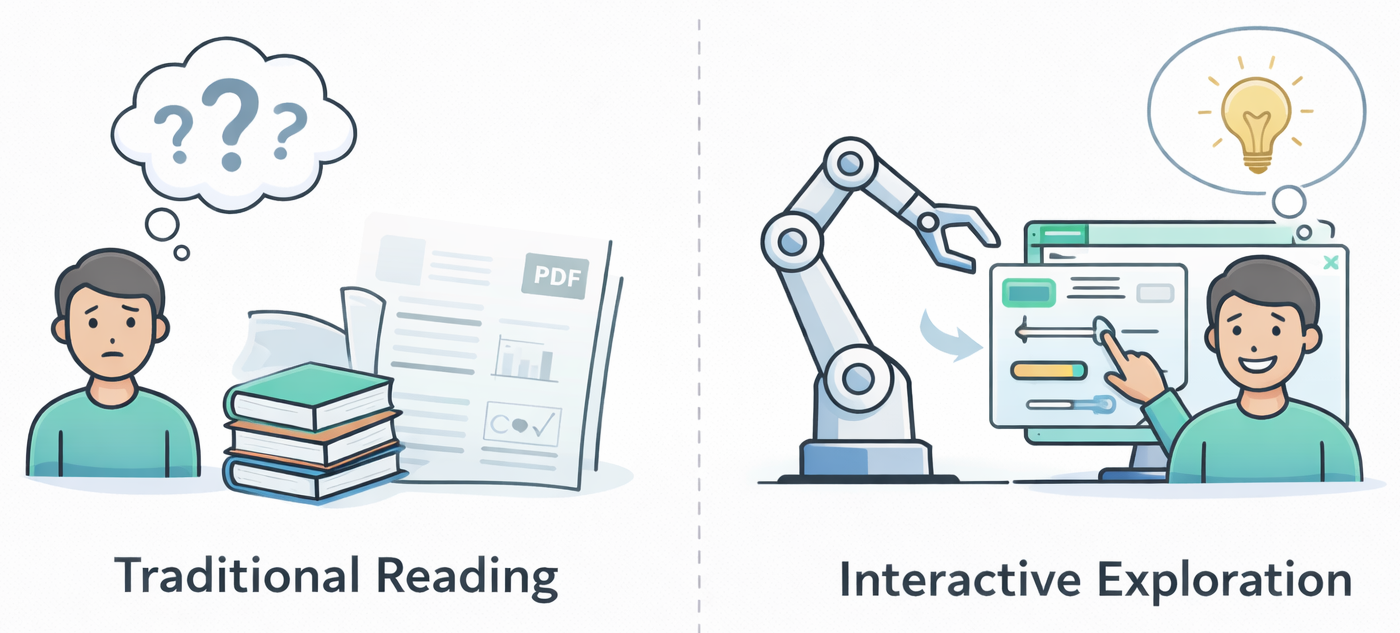}
  \vspace{-5mm}
  \caption{Traditional Reading vs. Interactive Exploration.Left: conventional paper comprehension relies on passive reading and mental simulation. Right: our approach enables \emph{interactive exploration}, where users manipulate controls in an executable interface and observe state changes to better understand the paper’s key mechanisms and dynamics.}
  \label{fig:setting_trad_vs_interactive}
  \vspace{-3mm}
\end{figure}

\begin{figure*}[t!]
\centering
\includegraphics[width=1\linewidth]{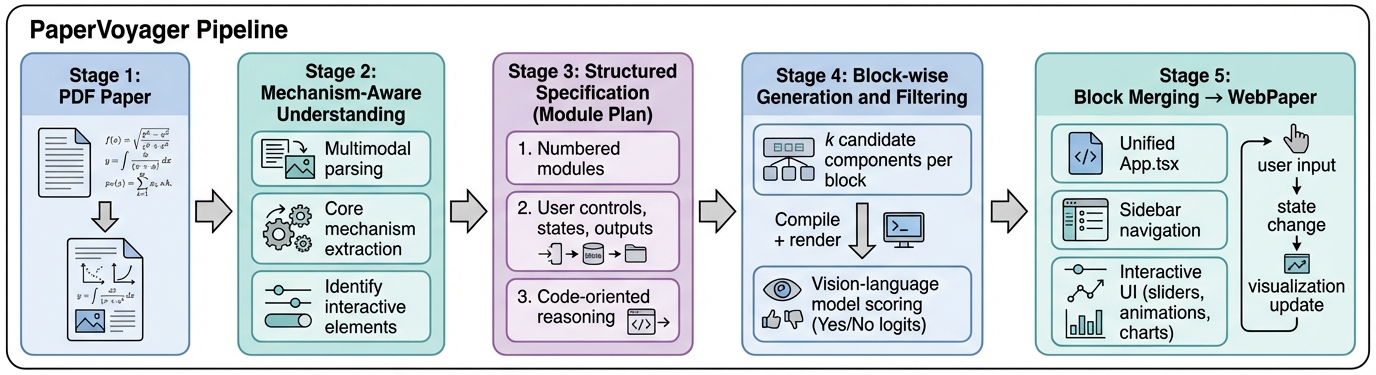}
\vspace{-5mm}
\caption{\textbf{Overview of the PaperVoyager pipeline.} Starting from a static PDF paper, the agent performs multimodal document parsing, identifies core mechanisms suitable for interaction, designs a structured generation specification, and synthesizes code via an LLM to produce an executable \emph{WebPaper}.}
\label{fig:overall}
\end{figure*}

To address these limitations, we present PaperVoyager, a Paper-to-Interactive-System Agent that transforms research papers from static text into interactive and executable web systems. This allows users to understand core mechanisms and dynamics through direct interaction. Given a research paper in PDF format, PaperVoyager operates in an end-to-end manner without human intervention. It performs three main steps: paper understanding, system modeling, and interactive webpage synthesis. First, the agent analyzes the paper to identify its main ideas and key mechanisms. It then determines which components can be explored interactively, such as algorithmic procedures, state evolution, or relationships between system components. Based on this analysis, PaperVoyager constructs a structured generation specification that defines the interactive modules, page layout, and user interaction logic. This specification is then used to guide a large language model to generate a complete interactive web system. Unlike traditional paper reading, which relies heavily on the reader’s mental simulation, PaperVoyager directly produces an executable interactive environment. Users can manipulate inputs, observe state transitions, and verify their understanding through interaction. In this way, paper comprehension shifts from passive reading to active exploration.

Evaluating the ability of models to synthesize interactive systems from research papers remains a significant challenge. Existing benchmarks for document understanding or webpage generation mainly focus on static content correctness or stepwise generation processes, which fail to capture the quality of interaction design and behavioral fidelity in executable systems~\citep{lu2025webgen,mialon2023gaia,yao2022webshop,jimenez2023swe}. To address this gap, we introduce a new benchmark consisting of 19 representative research papers, each paired with an expert-authored interactive web system whose source code serves as the ground-truth reference. The benchmark evaluates generated systems across three aspects: system structure, interactivity, and behavioral consistency with the source paper. Using this benchmark, we evaluate several existing large language models and find that although they perform reasonably well on isolated subtasks such as paper understanding or code generation, they struggle with the full process from paper understanding to system modeling and interactive synthesis. Common issues include coarse interaction design, incomplete system structures, and inconsistencies between system behavior and the mechanisms described in the paper. Motivated by these findings, we propose PaperVoyager, which explicitly models core mechanisms and interactive elements in research papers and incorporates structured constraints together with multi-stage coordinated reasoning during generation. Experiments show that PaperVoyager substantially outperforms existing large language models on our benchmark, demonstrating its effectiveness for end-to-end paper-to-interactive-system synthesis.

\noindent Our work makes the following key contributions:
\begin{itemize}[leftmargin=*, itemsep=0pt, topsep=0pt, parsep=0pt]
  \item We curate and annotate a new evaluation dataset for paper-to-interactive-system synthesis, where each research paper is paired with an expert-designed interactive web system as a ground-truth reference.
  \item We conduct systematic evaluations of representative, mainstream VLMs on this benchmark, providing a comprehensive analysis of their end-to-end capabilities and common failure modes in interactive system synthesis.
  \item We propose \textit{PaperVoyager}, a new approach that improves end-to-end paper-to-interactive-system generation, achieving substantial performance gains over strong baselines on the proposed benchmark.
\end{itemize}

\section{Related Work}

\subsection{Benchmarks for Coding LLMs.}
 
Numerous benchmarks have been proposed to evaluate the coding capabilities of large language models. Early evaluations mainly focus on well-specified programming tasks, where models generate functions or short programs that satisfy given test cases~\citep{hendrycks2021measuring,chen2021evaluating,austin2021program}. These datasets are typically collected from natural-language programming queries~\citep{zhang2024naturalcodebench}, competitive programming tasks~\citep{jain2024livecodebench}, automatically synthesized problems~\citep{zhuo2024bigcodebench}, and expert-designed evaluations~\citep{muennighoff2023octopack}. While effective for measuring code correctness, these benchmarks mainly evaluate function-level generation on relatively small, self-contained programs. More recent benchmarks extend evaluation to repository-level tasks that better reflect real-world development. For example, SWE-bench~\citep{jimenez2023swe,yang2024swe} and SWE-Lancer~\citep{miserendino2025swe} are built from real software repositories and issue reports, requiring models to implement fixes or features within existing codebases. These tasks involve bug fixing~\citep{jimenez2023swe,yang2024swe,aleithan2024swe}, code completion~\citep{liu2023repobench,zhang2023repocoder}, and multi-file patch generation~\citep{miserendino2025swe}. 

\subsection{Web Generation Agents} 

Recent advances in LLMs have enabled systems that assist or automate software and web development. Agent-style frameworks such as OpenHands~\citep{wang2024openhands}, SWE-agent~\citep{yang2024sweagent}, and Aider~\citep{aiderai2024aider} allow models to interact with tools and execution environments to iteratively generate and refine code. Development assistants including Cursor~\citep{Cursor2024cursor}, GitHub Copilot~\citep{GitHub2024copilot}, and Devin~\citep{Wu2024devin} further demonstrate the potential of LLM-driven programming systems. Beyond general code generation, recent work explores integrating visual information into coding tasks. In particular, VLM have been studied for generating webpage code from visual inputs, such as reconstructing HTML from webpage screenshots~\citep{guo2024iwbenchevaluatinglargemultimodal,si2025design2codebenchmarkingmultimodalcode,yun2024web2codelargescalewebpagetocodedataset,beltramelli2017pix2codegeneratingcodegraphical}. Other studies evaluate models on implementing interactive elements or solving web development tasks through predefined pipelines~\citep{xiao2025interaction2codebenchmarkingmllmbasedinteractive,xiao2025designbenchcomprehensivebenchmarkmllmbased,xu2025webbenchllmcodebenchmark}. However, these settings often focus on single-file HTML generation or rigid pipelines, which are better suited for evaluating multimodal models rather than interactive web systems. In contrast, our work introduces a benchmark for dynamic webpage generation, focusing on the ability to construct interactive web systems from scratch.

\begin{figure}[t!]
\small
\begin{center}
\fbox{
\parbox{0.95\linewidth}{

You are an expert developer that converts research papers into interactive web demos. Your task is to generate an interactive educational web application that helps users understand the core mechanisms of a research paper. Before generating code, you must:

(1). Identify the key mechanisms in the paper that should be visualized.

(2). Design interactive modules that allow users to explore these mechanisms.

(3). Specify the user controls and visual outputs.

Then generate a complete single-page web application implemented using React and TypeScript.
}
}
\end{center}
\vspace{-3mm}
\caption{Prompt used by PaperVoyager to generate interactive web systems.}
\vspace{-3mm}
\end{figure}

\section{PaperVoyager}

We introduce \emph{PaperVoyager}, an autonomous agent that transforms research
papers into executable interactive web systems. Given a research paper in PDF
format, PaperVoyager parses the document, identifies its core mechanisms, and
synthesizes an interactive web application that allows users to explore the
paper through direct interaction. Unlike traditional document assistants that
mainly convert papers into static artifacts, PaperVoyager focuses on
constructing runnable systems that expose the state, dynamics, and mechanisms
described in technical papers.

After parsing the paper, PaperVoyager does not directly generate code. Instead,
it first prompts the model to reason explicitly about which mechanisms should be
exposed through interactive visualization.  The model identifies key presentation points suitable for dynamic exploration, including algorithm execution steps, parameter-driven behaviors, and state transitions, and constructs a structured specification that enumerates the
interactive modules, user controls, and expected visual outputs. This specification, which includes a numbered \emph{Module Plan}, acts as a code-oriented chain-of-thought that decomposes the paper into a set of self-contained interactive modules before any code is written. Further details on multimodal parsing, block-level generation, candidate filtering, and block merging are provided in Appendix A.

\section{Benchmark for PaperVoyager}\label{sec:benchmark}

To evaluate the ability of LLM to generate \textit{interactive, dynamic web applications}, we construct a benchmark consisting of \textit{19 topic-conditioned Web tasks}. Each task requires generating a self-contained single-page web application that teaches a specific scientific concept through interactive visualizations. Each Web instance includes a landing page and multiple interactive learning modules (e.g., parameterized simulations, animated state transitions, and real-time visualizations). Users interact with the system through UI controls such as sliders, buttons, or dropdown menus to explore how the underlying mechanisms evolve dynamically. In our evaluation, each generated Web instance is treated as an independent website with a fixed start page.

\subsection{Paper Selection}

The benchmark spans seven scientific domains, including computer science,
mathematics, and physics. It covers a wide range of visualization complexity,
ranging from discrete computational processes to continuous dynamical systems
governed by differential equations. To construct the benchmark, we
curate 19 topics rooted in landmark papers and foundational algorithms across these domains. Table~\ref{tab:benchmark_topics} enumerates the full topic set alongside their originating works and domains. Each topic is selected according to a single criterion: the core mechanism
should admit a natural interactive representation, such as parameter sliders,
animated state transitions, or real-time canvas rendering.

\begin{table}[t!]
\centering
\small
\setlength{\tabcolsep}{3pt}
\resizebox{\columnwidth}{!}{
\begin{tabular}{@{}lll@{}}
\toprule
\textbf{Abbrev.} & \textbf{Topic} & \textbf{Domain} \\
\midrule
Alg-DP      & Dynamic Programming        & Algorithms        \\
Alg-GP      & Graph Pathfinding          & Algorithms        \\
Alg-SR      & Sorting Algorithms         & Algorithms        \\
DS-BT       & Balanced BSTs              & Data Structures   \\
DS-HM       & Hash Maps / Cuckoo Hashing & Data Structures   \\
Dist-Raft   & Raft Consensus             & Distributed Sys.  \\
Math-Lorenz & Lorenz Attractor           & Mathematics       \\
Math-FFT    & Fourier Series / FFT       & Mathematics       \\
Math-Eig    & Eigendecomposition         & Mathematics       \\
Math-MC     & Monte Carlo Estimation     & Mathematics       \\
ML-GD       & Gradient Descent           & Machine Learning  \\
ML-KM       & K-Means Clustering         & Machine Learning  \\
ML-NNV      & Neural Net Backprop        & Machine Learning  \\
Phys-CFD    & 2D Fluid Simulation        & Physics           \\
Phys-Orbit  & N-body Gravity             & Physics           \\
Phys-Opt    & Optics \& Ray Tracing      & Physics           \\
Phys-Therm  & Thermodynamics             & Physics           \\
Sys-Sched   & CPU Scheduling             & Systems           \\
Sys-VM      & Virtual Memory \& Paging   & Systems           \\
\bottomrule
\end{tabular}
}
\caption{Benchmark topics and domain categories. The complete list of papers and detailed metadata are provided in the Appendix.}
\label{tab:benchmark_topics}
\end{table}


\subsection{Data Construction}

Constructing high-quality, executable prompt specifications and reference implementations from research papers requires bridging dense academic notation with concrete UI/UX affordances. To establish a rigorous ground truth for our benchmark, we adopt a meticulous human-in-the-loop construction pipeline comprising three main stages.

\paragraph{Stage 1: Expert Mechanism Extraction.}
For each selected topic, a PhD-level domain expert meticulously studies the originating research paper to identify the core mechanisms that benefit most from interactive exploration. The expert determines the critical abstract concepts (e.g., algorithmic steps, physical dynamics, or system state transitions) and maps them to concrete user interactions. This phase produces a detailed interaction blueprint specifying the required interactive modules, key manipulative parameters (e.g., UI controls such as sliders and buttons), and the expected visual feedback that confirms conceptual understanding.

\paragraph{Stage 2: Synthesis Web Code.}
Translating conceptual blueprints into complex multi-module web applications is highly non-trivial and requires substantial engineering effort. Based on the interaction blueprint, human experts first define the target functionality, interaction behavior, and visual requirements for each module. Code-specialized language models are then used only as engineering assistants to accelerate the initial implementation of React and TypeScript components. The resulting drafts are manually inspected, revised, and standardized by human experts to ensure that they faithfully implement the intended paper mechanisms and interactive behaviors.

Importantly, these reference implementations are used only to define the desired functional behavior of each task, rather than to provide code-level or style-level targets for model evaluation. Generated systems are evaluated according to functionality-oriented criteria, including checklist coverage and interaction reliability, rather than similarity to any particular implementation produced during benchmark construction.

\paragraph{Stage 3: Verification and Refinement.}
To ensure the highest level of pedagogical fidelity and technical correctness, the LLM-generated code is not used as-is. In this final stage, domain experts conduct rigorous manual verification and refinement. They systematically test the generated interactive systems, debug logic errors, and refine UI/UX details. Most importantly, experts carefully adjust the underlying state transition logic to guarantee strict behavioral consistency with the mathematical or algorithmic definitions in the original paper. This manual polishing process yields the artifact-level source code that serves as the gold-standard ground truth for our benchmark evaluation.

\begin{table*}[t!]
\centering
\setlength{\tabcolsep}{1.6mm}{
\resizebox{\textwidth}{!}{%
\begin{tabular}{@{}lcccccccccc@{}}
\toprule \toprule
Metric &
Alg-DP & Alg-GP & Alg-SR &
DS-BT & DS-HM &
Dist-Raft &
Math-Lorenz & Math-FFT & Math-Eig & Math-MC \\
\midrule
GPT5.2 &
39.9 & 41.7 & 52.6 &
33.9 & 46.3 &
53.0 &
39.2 & 77.3 & 72.4 & 53.5 \\
MiniMax &
8.5 & 12.4 & 23.1 &
56.1 & 43.5 &
27.3 & 73.3 &
23.5 & 30.3 &
67.6 \\
Qwen-Max &
8.0 & 12.4 & 42.4 &
28.9 & 29.9 &
27.3 & 69.2 &
27.4 & 44.3 &
79.0 \\
Kimi-K2 &
54.8 & 40.6 & 53.9 &
29.8 & 54.7 &
55.3 &
50.4 & 84.3 & 69.1 & 62.0 \\

Gemini-3-Pro &
51.7 & 80.7 & 55.0 &
47.3 & 41.7 &
50.0 &
30.7 & 32.0 & 55.0 & 45.0 \\

\midrule
\textbf{PaperVoyager (Ours)} &
\textbf{93.0} & 75.0 & 53.3 &
\textbf{52.7} & \textbf{72.0} &
\textbf{82.0} &
\textbf{89.0} & \textbf{92.3} & \textbf{85.0} & \textbf{93.7} \\

\midrule
\midrule
Metric &
ML-GD & ML-KM & ML-NNV &
Phys-CFD & Phys-Orbit & Phys-Opt & Phys-Therm &
Sys-Sched & Sys-VM &
Overall \\
\midrule
GPT5.2 &
24.9 & 30.6 & 29.4 &
53.8 & 41.4 & 71.9 & 40.3 &
52.4 & 45.8 &
46.2 \\
MiniMax &
79.0 & 83.0 & 83.5 &
81.3 & 83.9 & 59.5 & 84.8 &
71.0 & 26.5 &
55.0 \\
Qwen-Max &
29.5 & 73.5 & 0.0 &
62.4 & 32.7 & 55.3 & 53.3 &
54.5 & 30.4 &
39.2 \\
Kimi-K2 &
38.8 & 74.3 & 29.3 &
63.5 & 84.5 & 38.1 & 0.0 &
62.6 & 52.6 &
54.2 \\


Gemini-3-Pro   &
84.0 & 51.7 & \textbf{84.0} &
73.3 & 55.0 & 81.3 & 57.3 &
35.0 & 85.0 &
57.5 \\

\midrule
\textbf{PaperVoyager (Ours)} &
\textbf{92.7} & \textbf{95.0} & 83.0 &
\textbf{88.0} & \textbf{89.7} & \textbf{85.0} & \textbf{92.3} &
\textbf{84.0} & \textbf{92.0} &
\textbf{84.4} \\

\bottomrule \bottomrule
\end{tabular}
 }}
\vspace{-2mm}
\caption{
Task success rates across 19 representative tasks. For all baseline models and PaperVoyager variants, the final scores are computed based on Checklist Matching Evaluation and Interactive Exploration Evaluation, with weights of 60\% and 40\%, respectively.
}
\label{tab:task_breakdown_final}
\end{table*}

\section{Experiment}

\subsection{Implementation Details}


The generated application is packaged into a deployable website using a standard web build toolchain, yielding the final WebPaper instance. All interactions are executed in a reproducible browser environment (Chromium via Playwright) with a fixed viewport of $1024 \times 768$ pixels, ensuring consistent screenshot-based observations across runs. During task execution and probing, the system records complete interaction trajectories, including stepwise actions and screenshots, which are later consumed by both deterministic and model-based evaluators.

\subsection{Evaluation Methods}

We evaluate generated web systems using two complementary methods that measure both design compliance and interactive functionality.

\paragraph{Checklist Matching Evaluation.}
We first measure whether the generated system implements the intended interactive design specified by our annotation checklist. The checklist is manually constructed and contains a list of required visualization modules and interactive elements that should be used to present key concepts in each paper. Each item specifies which content should be demonstrated through visualization or interaction. To enable automatic evaluation, we require the code generation model to provide explicit module-level comments describing the purpose of each interactive component. After generation, we use Gemini to extract all module descriptions from the source code. These extracted module comments are then matched against the checklist items to determine whether the required modules are implemented. The final completion rate is computed as the proportion of checklist items successfully matched by generated modules.

\begin{figure}[t!]
\centering
\includegraphics[width=1\linewidth]{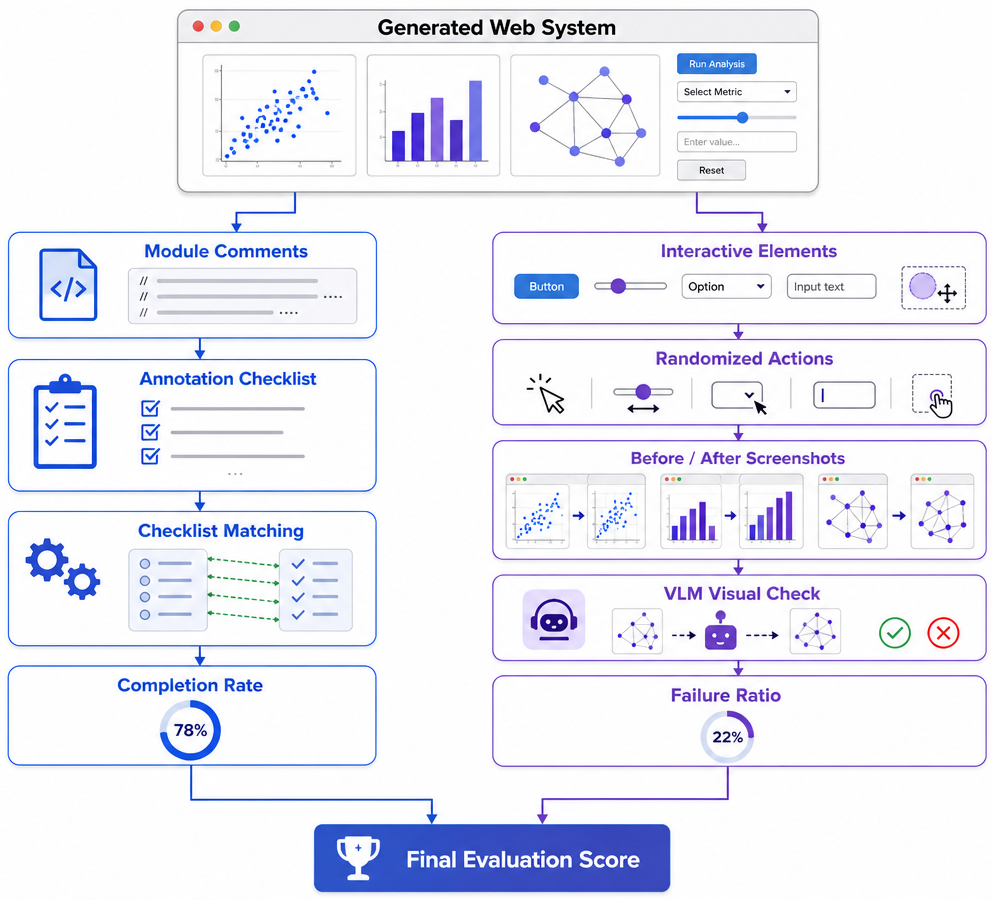}
\vspace{-7mm}
\caption{Two-branch evaluation protocol for generated interactive web systems, combining checklist completion and interaction failure analysis into a final quality score.}
\label{fig:overall2}
\vspace{-3mm}
\end{figure}

\paragraph{Interactive Exploration Evaluation.}
While checklist matching measures whether the required interactive modules are present, it does not verify whether these modules actually respond to user actions. We therefore introduce an automatic interaction-based evaluation that tests the functional responsiveness of the generated webpage. Given the full webpage source code, we first identify interactive DOM elements, including buttons, sliders, dropdown menus, input fields, and draggable or clickable canvas regions. For each detected element, the evaluator performs a predefined random action sampled from its supported interaction type, such as clicking a button, moving a slider to a random value, selecting an option from a dropdown menu, typing into an input box, or dragging a canvas object.

Before and after each interaction, we capture webpage screenshots and compare the resulting visual states. A lightweight VLM agent is then asked to determine whether the interaction produces a meaningful and visible change that is consistent with the intended module behavior. For example, a valid interaction may update a visualization, change a parameter value, highlight an algorithm step, move an object, or reveal new explanatory content. If no observable visual change is detected after the interaction, or if the page becomes broken or visually inconsistent, the corresponding component is marked as a failed interaction. We report the failure ratio over all tested interactive elements as a measure of interactive reliability, where a lower failure ratio indicates a more functional generated system.

\subsection{Main Results}

Table 2 presents the task success rates across 19 representative tasks. Overall, PaperVoyager achieves the best performance with an average SR of 80.7\%, outperforming all baseline models. The strongest baseline is Qwen-Max, with an overall score of 80.2\%. Kimi 2.5 and Gemini-3-Flash both achieve 77.3\%, Minimax reaches 73.6\%, and GPT5.2 performs the worst with 68.1\%.

Across different task categories, PaperVoyager shows consistently strong performance. In algorithm and data structure tasks, including Alg-DP, Alg-GP, and DS-HM, our method achieves competitive or leading results. In particular, Alg-GP reaches 90.4\%, indicating that the generated interfaces effectively capture algorithmic dynamics such as graph traversal and pathfinding processes. In machine learning tasks, including ML-GD, ML-KM, and ML-NNV, PaperVoyager achieves 85.3\%, 83.3\%, and 84.5\%, respectively. This suggests that the generated systems successfully expose optimization processes and model behaviors through interactive visualizations.



\begin{table}[t]
\centering
\setlength{\tabcolsep}{2.2mm}
\resizebox{\columnwidth}{!}{%
\begin{tabular}{@{}ccccccccc@{}}
\toprule
Metric & Alg & DS & Dist & Math & ML & Phys & Sys & Overall \\
\midrule
k=1 & 63.5 & 73.2 & 74.3 & 78.5 & 69.6 & 77.9 & 62.3 & 73.4 \\
k=2 & 66.0 & 73.9 & 81.1 & 79.3 & 83.6 & 78.8 & 64.4 & 76.1 \\
k=3 & 66.3 & 74.7 & 81.1 & 80.3 & 85.4 & 78.5 & 79.7 & 78.1 \\
k=4 & 67.3 & 75.8 & 81.1 & 80.3 & 86.0 & 78.5 & 79.8 & 78.4 \\
k=6 & 79.5 & 75.2 & 81.1 & 80.5 & 87.0 & 78.7 & 80.0 & 80.4 \\
\midrule
k=5  & 79.1 & 74.9 & 81.1 & 80.3 & 86.7 & 78.5 & 79.7 & 80.1 \\
\bottomrule
\end{tabular}%
}
\vspace{-2mm}
\caption{
Task completion rates under different numbers of generation attempts k. }
\label{tab:task_breakdown_grouped_avg}
\vspace{-5mm}
\end{table}


\begin{figure*}[t!]
    \centering

    \begin{subfigure}[b]{0.24\textwidth}
        \centering
        \includegraphics[width=\textwidth]{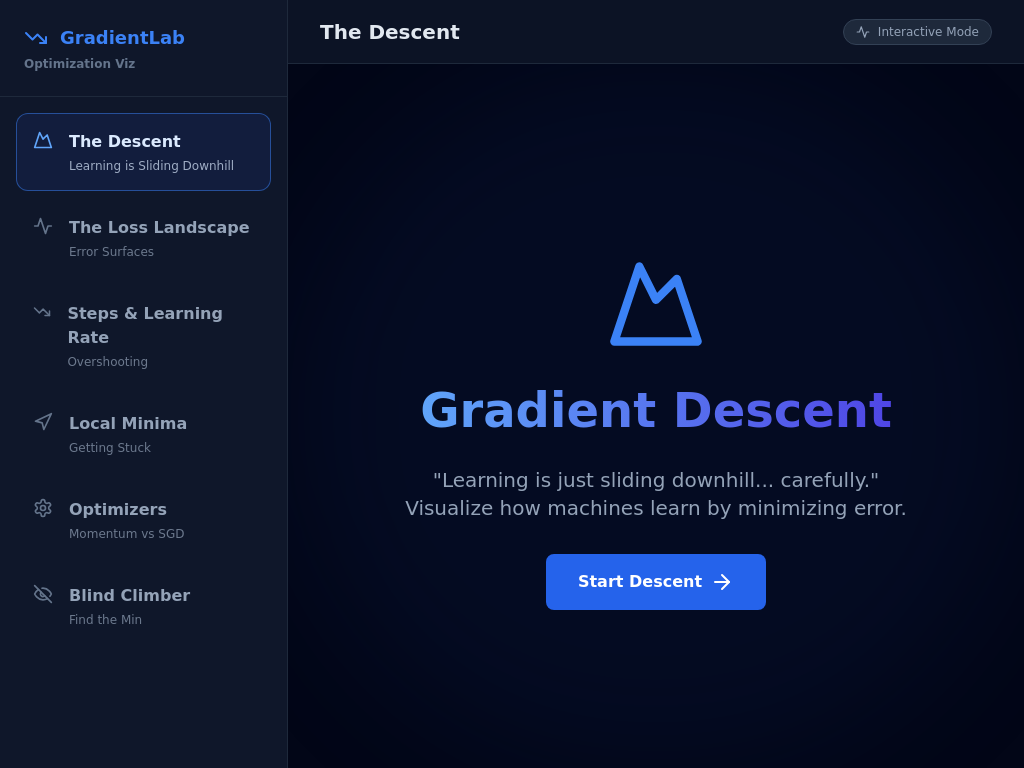}
        \label{fig:gd_g_landing}
    \end{subfigure}
    \begin{subfigure}[b]{0.24\textwidth}
        \centering
        \includegraphics[width=\textwidth]{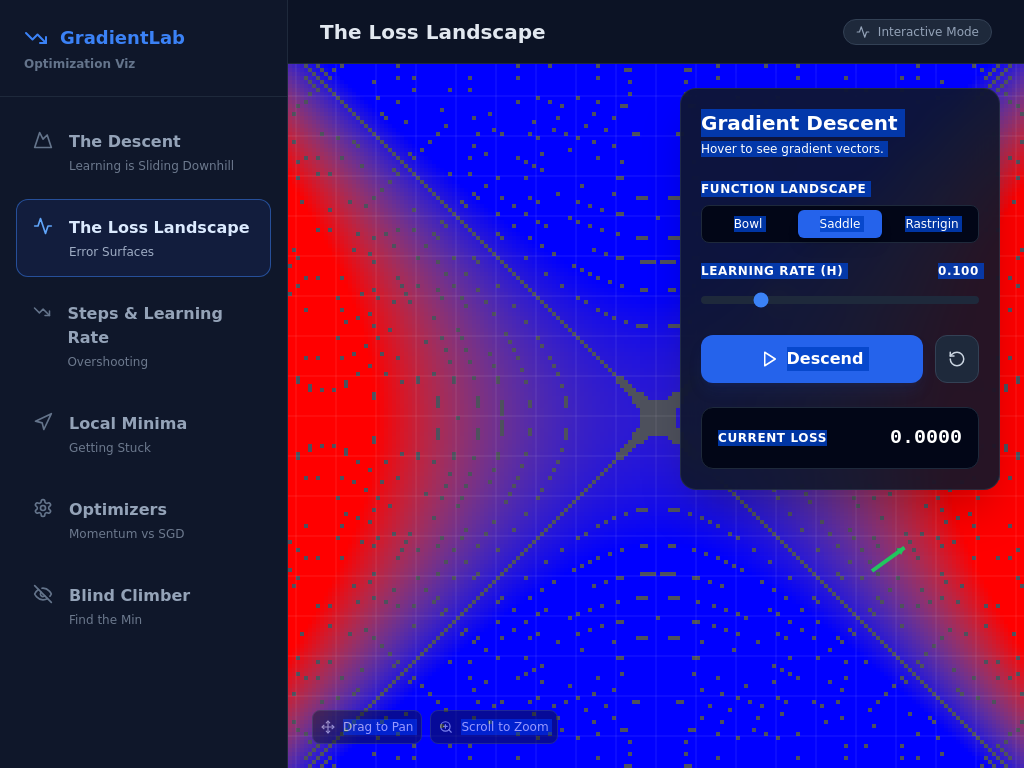}
        \label{fig:gd_g_func}
    \end{subfigure}
    \begin{subfigure}[b]{0.24\textwidth}
        \centering
        \includegraphics[width=\textwidth]{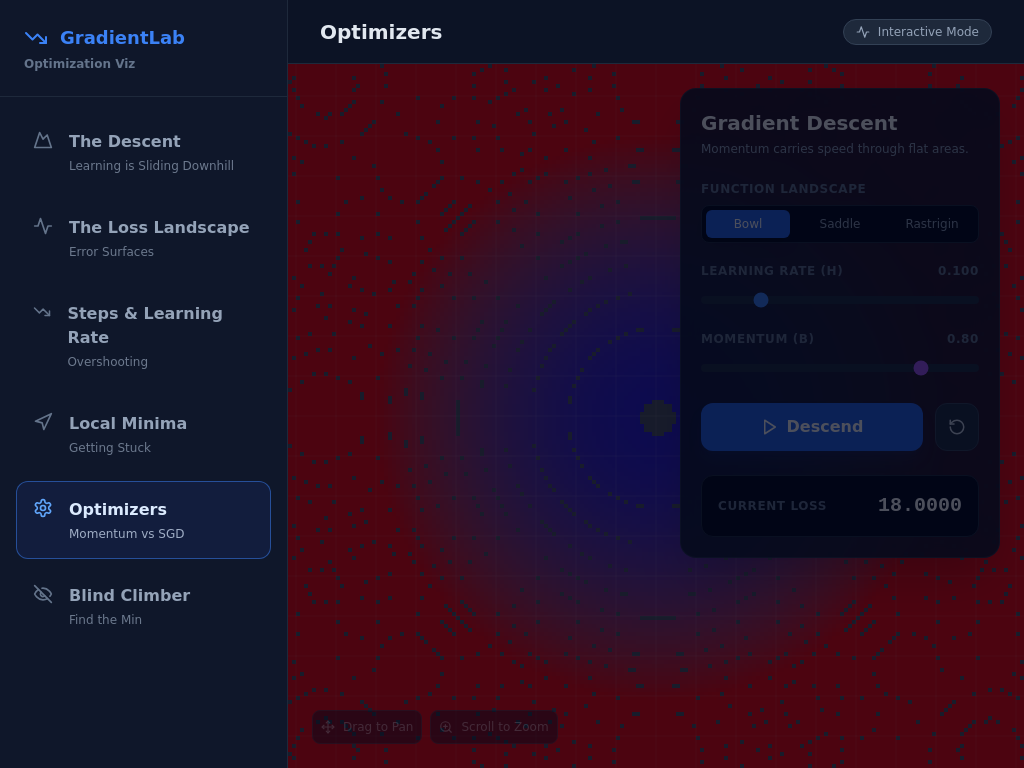}
        \label{fig:gd_g_opt}
    \end{subfigure}
    \begin{subfigure}[b]{0.24\textwidth}
        \centering
        \includegraphics[width=\textwidth]{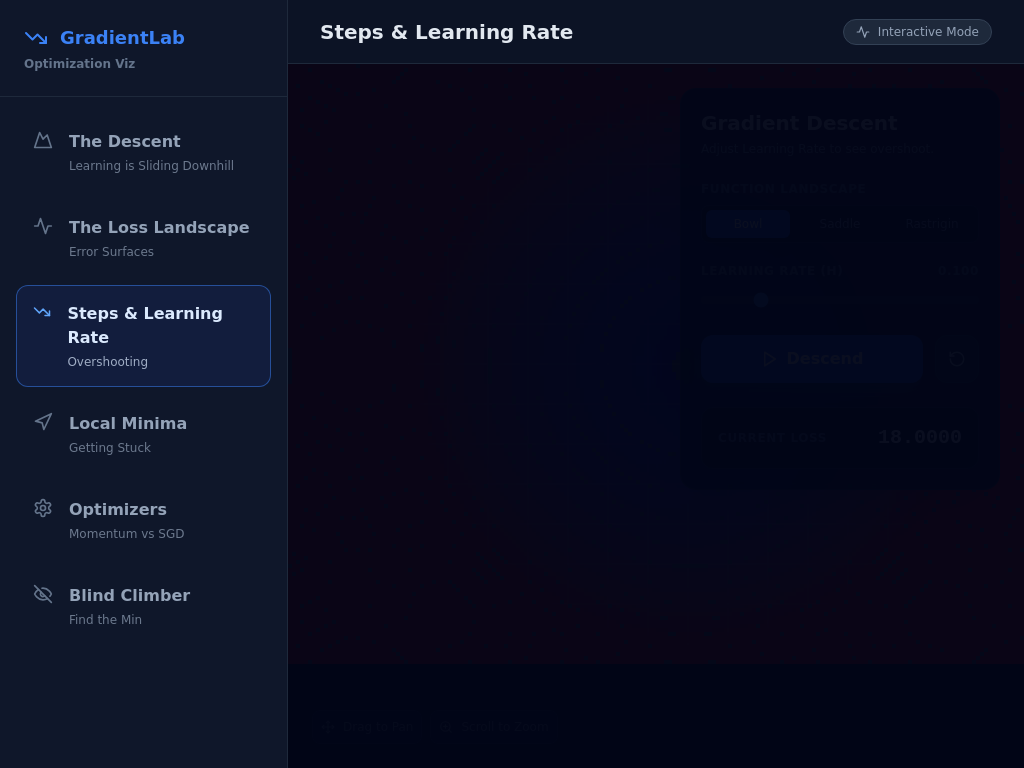}
        \label{fig:gd_g_lr}
    \end{subfigure}

    \vspace{-4mm}

    \begin{subfigure}[b]{0.24\textwidth}
        \centering
        \includegraphics[width=\textwidth]{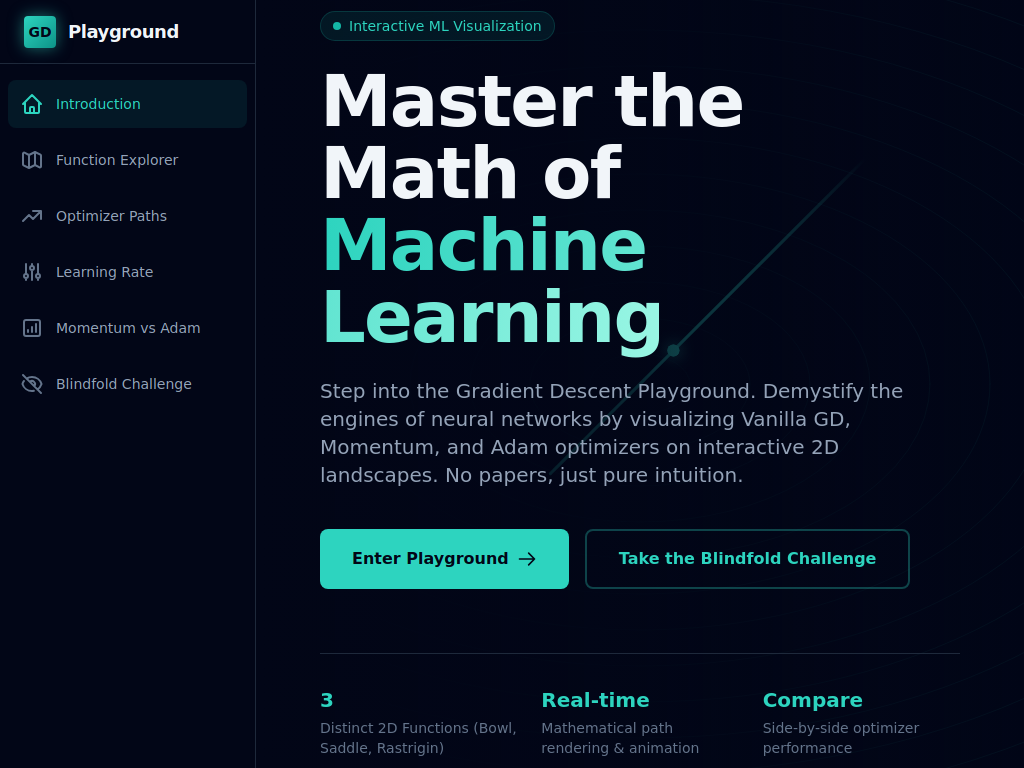}
        \label{fig:gd_pv_landing}
    \end{subfigure}
    \begin{subfigure}[b]{0.24\textwidth}
        \centering
        \includegraphics[width=\textwidth]{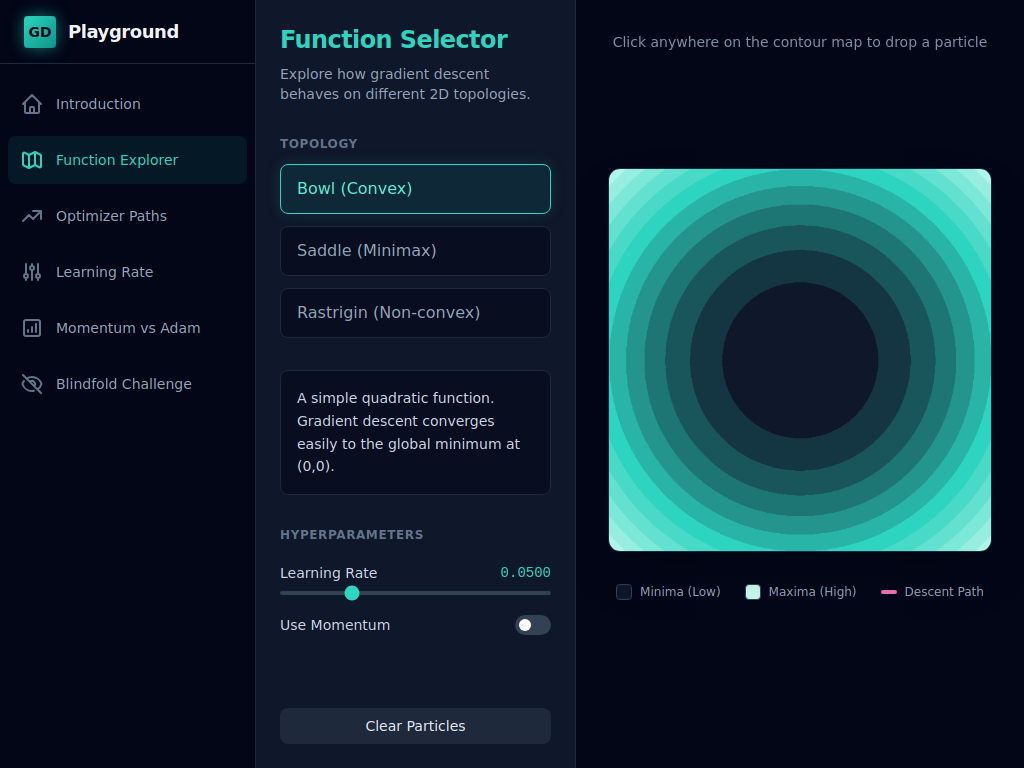}
        \label{fig:gd_pv_func}
    \end{subfigure}
    \begin{subfigure}[b]{0.24\textwidth}
        \centering
        \includegraphics[width=\textwidth]{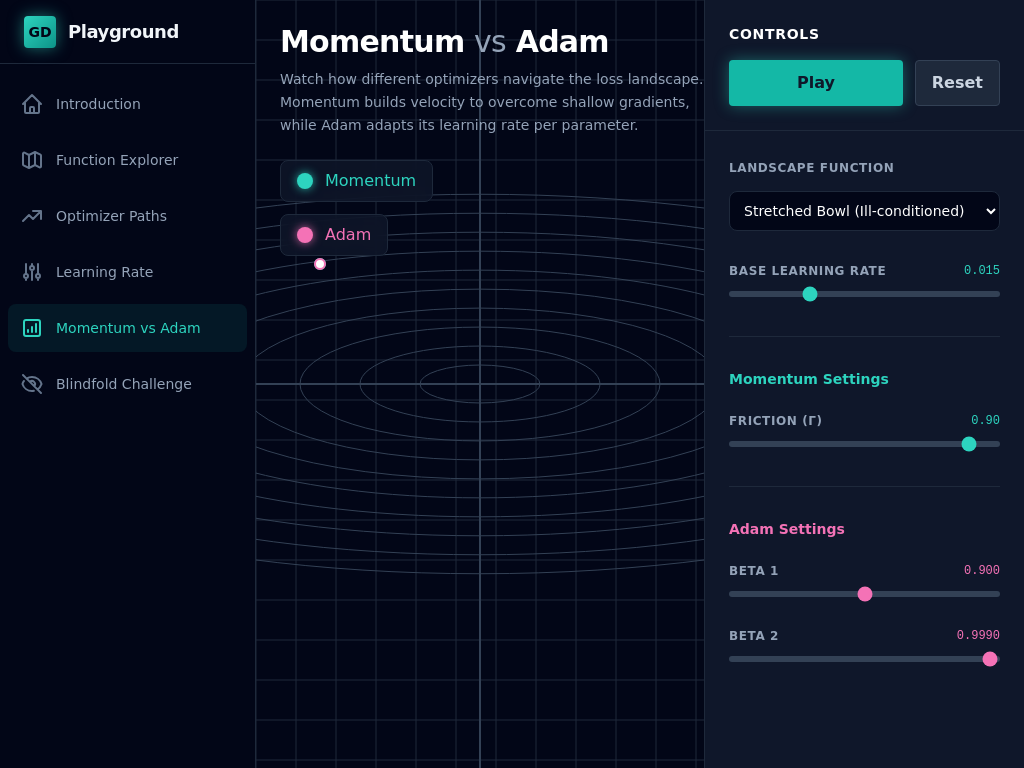}
        \label{fig:gd_pv_opt}
    \end{subfigure}
    \begin{subfigure}[b]{0.24\textwidth}
        \centering
        \includegraphics[width=\textwidth]{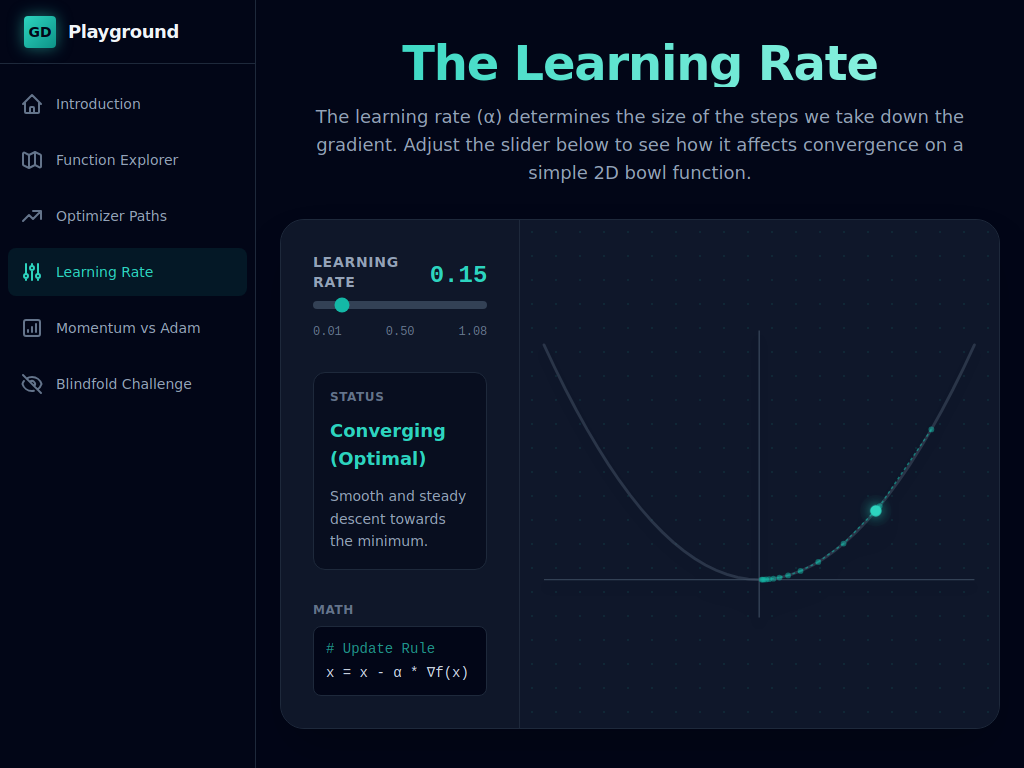}
        \label{fig:gd_pv_lr}
    \end{subfigure}

    \vspace{-5mm}

    \caption{Case study on \textit{ML Gradient Descent}.
    Row~1 shows the single-shot baseline with Gemini-3-Pro, and Row~2 shows PaperVoyager.  }
    \label{fig:gd_casestudy}
\end{figure*}


\subsection{Ablation Study}

To further understand the role of the VLM-based evaluator in improving generation quality, we conduct a set of ablation studies that analyze how the evaluator leverages internal visual-language signals to assess generated webpages.  We design two complementary ablation settings to study this mechanism. The first investigates how the amount of trajectory information (i.e., the number of screenshots provided to the evaluator) affects the reliability of the evaluation signal. The second examines the effect of performing multiple independent generations and selecting the highest-scoring result. Together, these experiments reveal how richer trajectory observations and a larger candidate pool enable the VLM evaluator to better identify high-quality WebPaper implementations.


\paragraph{Effect of Multiple Generation.}

Table~\ref{tab:task_breakdown_grouped_avg} reports the effect of increasing the number of generation attempts.  In this setting, the evaluator is provided with the full interaction trajectory of the generated webpages.  For each task, the system performs multiple independent generations, and the VLM evaluator assigns a score to each generated result.  The candidate with the highest evaluation score is then selected as the final output for overall evaluation. Increasing the number of generation attempts generally improves the final performance, since the evaluator can select higher-quality results from a larger candidate pool. This strategy effectively reduces the impact of occasional generation failures and improves the robustness of the system. Overall, combining multiple generations with VLM-based selection provides a simple yet effective mechanism for improving the quality of generated WebPaper systems.

\begin{table}[t]
\centering
\setlength{\tabcolsep}{2.2mm}
\resizebox{\columnwidth}{!}{%
\begin{tabular}{@{}ccccccccc@{}}
 \toprule
  Metric & Alg & DS & Dist & Math & ML & Phys & Sys &
  Overall \\
  \midrule
  screen=0 & 62.5 & 44.5 & 50.0 & 40.7 & 73.2 & 66.7 & 60.0
   & 57.5 \\
  screen=1 & 65.8 & 50.1 & 58.5 & 55.2 & 78.1 & 73.0 & 67.8
   & 65.3 \\
  screen=2 & 68.5 & 54.0 & 67.8 & 67.5 & 82.8 & 79.1 & 75.2
   & 72.4 \\
  screen=3 & 71.6 & 59.2 & 76.8 & 81.9 & 87.4 & 85.2 & 83.3
   & 79.9 \\
  screen=4 & 72.7 & 61.0 & 79.3 & 85.8 & 88.8 & 86.9 & 85.7
   & 82.1 \\
  screen=5 & 73.2 & 61.7 & 80.5 & 87.8 & 89.7 & 88.1 & 87.1
   & 83.2 \\
  \midrule
  Full      & 73.8 & 62.4 & 82.0 & 90.0 & 90.2 & 88.8 &
  88.0 & 84.4 \\
  \bottomrule
\end{tabular}%
}
\vspace{-2mm}
\caption{
Task completion rates with different numbers of trajectory screenshots used by the VLM evaluator. 
Here, ``screen'' denotes the number of screenshots sampled from the interaction trajectory. 
}
\label{tab:task_breakdown_grouped_avg_screen}
\vspace{-3mm}
\end{table}


\paragraph{Effect of Observation Length.}
Table~\ref{tab:task_breakdown_grouped_avg_screen} reports the task success rates when different numbers of trajectory screenshots are used for evaluation. Here, $k$ denotes the number of screenshots sampled from the interaction trajectory. When no trajectory information is provided ($k=0$), the evaluator relies on minimal visual evidence, leading to relatively low performance. As the number of screenshots increases, the evaluator obtains richer information about the webpage state and interaction process, resulting in steadily improved selection quality. The performance gradually improves from $57.5$ at $k=0$ to $79.9$ at $k=3$. When the full trajectory is provided, the evaluator achieves the best performance, reaching an overall TSR of $84.4$. These results indicate that richer trajectory observations enable the VLM evaluator to better understand the dynamic behavior of the generated webpages, leading to more reliable quality assessment and selection.

\subsection{Further Analysis}
 
\paragraph{Quantitative Analysis.}
We examine the \textit{ML Gradient Descent} topic to illustrate a failure mode of monolithic generation: the systematic omission of paper-specified components.  Although the baseline model is strong, with Gemini-3-Pro achieving a score of 84.0 among the highest across tasks, Figure~\ref{fig:gd_casestudy} shows that several specification details are lost when the entire application is generated in a single pass. The specification derived from the paper defines six modules, including a \emph{Momentum vs.\ Adam} optimizer comparison and a function landscape explorer with bowl, saddle, and Rastrigin surfaces. Under monolithic generation, the baseline replaces Adam with the simpler SGD (column3), likely because implementing Adam requires additional logic for adaptive learning rates and moment estimation. A similar simplification occurs in the function explorer (column2), where the baseline renders the loss surface but omits the hyperparameter control panel required by the specification. In contrast, PaperVoyager generates each module independently; when the optimizer comparison is generated as a dedicated block, the model has sufficient generation budget to implement Adam correctly, achieving higher specification adherence.

\begin{table}[t!]
\centering
\resizebox{0.7\columnwidth}{!}{
\begin{tabular}{lc}
\toprule
\textbf{Main reasons for Failure} & \textbf{Ratio} \\
\midrule
Prompt Misalignment    & 43.3\% \\
Visual Grounding Issue & 35.6\% \\
Hallucination          & 14.8\% \\
Navigation Stuck       &  6.3\% \\
\bottomrule
\end{tabular}
}
\caption{Distribution of main failure reasons.}
\label{tab:failure_reasonsxxx}
\vspace{-5mm}
\end{table}

\paragraph{Qualitative Analysis.}
To better understand the limitations of current generation systems, we analyze failure cases in our benchmark. Table~\ref{tab:failure_reasonsxxx} summarizes four main failure categories. The most common is prompt misalignment (43.3\%), where the generated webpage does not follow the specification derived from the source paper and prompt, often resulting in missing modules or incorrect functionality. For example, in the \textit{ML Gradient Descent} task, the specification requires an \emph{Adam vs.\ Momentum} comparison, but the baseline replaces Adam with the simpler SGD optimizer. The second category is visual grounding issues (35.6\%), where the interface structure is partially generated but visual elements fail to render correctly, often appearing as missing visualizations or placeholder content such as ``coming soon''. Hallucination accounts for 14.8\% of failures, where the system introduces components or behaviors not specified in the paper or prompt. Finally, navigation stuck (6.3\%) occurs when user interactions fail to trigger meaningful state changes in the webpage. We detect such cases by measuring the visual difference between screenshots before and after an interaction; near-zero differences indicate failed transitions. Overall, most failures stem from specification misalignment and rendering errors, highlighting the importance of accurate instruction following and reliable visual grounding in WebPage generation.

\begin{figure}[t!]
    \centering
    \includegraphics[width=\columnwidth]
    {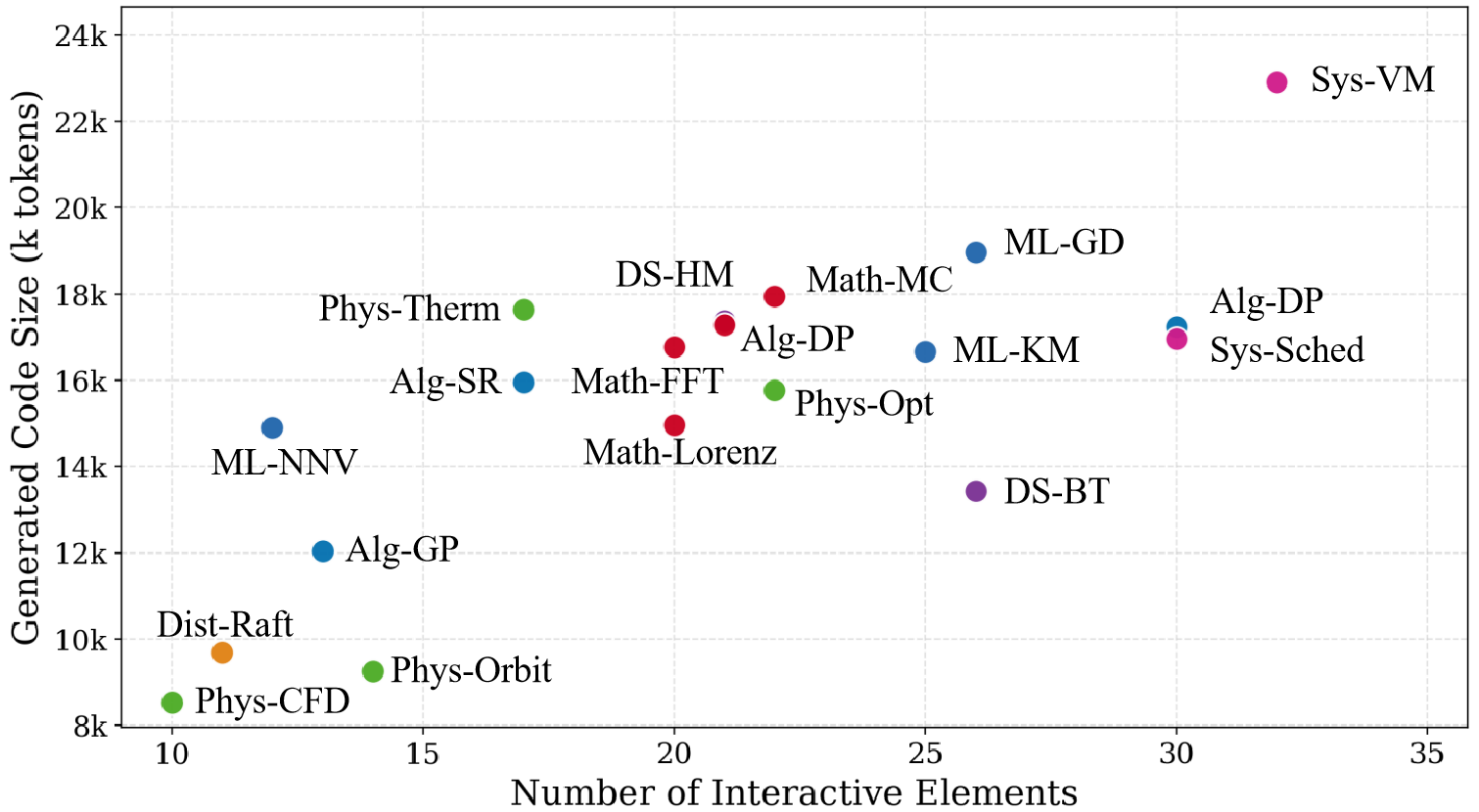}
    \caption{The relationship between the number of interactive elements and the generated code size (in tokens) across different scientific domains.}
    \label{fig:blocks_vs_tokens}
    \vspace{-3mm}
\end{figure}

\paragraph{Complexity Across Domains.}

Figure~\ref{fig:blocks_vs_tokens} shows the relationship between the number of interactive elements and the generated code size (tokens) for each WebDemo instance. Each point represents a task in our benchmark, with colors indicating different scientific domains. In general, tasks with more interactive elements require larger code sizes, reflecting the increased complexity of synthesizing multi-module interactive systems.

However, the complexity varies systematically across domains. Tasks from natural sciences, such as physics simulations (e.g., Phys.~CFD or Phys.~Orbit), typically require fewer interactive elements (often fewer than 15) and relatively smaller token budgets. Many physical phenomena have intuitive visual representations, such as trajectories or dynamic motion, allowing their mechanisms to be captured with relatively compact interfaces. In contrast, tasks derived from computer systems and complex algorithms often require significantly more interactive elements and larger code sizes. Concepts such as virtual memory (Sys.~VMem), CPU scheduling (Sys.~CPU), or dynamic programming (Alg.~DP) are inherently abstract and require exposing intermediate states and multiple sub-mechanisms, resulting in more modular interfaces and higher generation complexity. These results suggest that the complexity of WebDemo generation is influenced not only by interface design but also by the nature of the underlying domain: observable physical processes tend to yield more compact systems, while

\section{Conclusion}
 
In this work, we introduce PaperVoyager, an autonomous agent that transforms static research papers into executable, interactive web systems without human intervention. Unlike existing document agents that merely convert papers into static formats (such as summaries or slides), PaperVoyager explicitly models the core mechanisms, state transitions, and interaction logic to enable active exploration of complex technical concepts. To rigorously evaluate this novel task, we curated a new benchmark comprising 19 representative research papers across diverse scientific domains, each paired with an expert-authored interactive web system as the ground-truth reference. Through systematic evaluation of representative VLMs, we demonstrate that existing models struggle with the full process of end-to-end interactive synthesis. In contrast, our proposed PaperVoyager, utilizing structured specification design and a VLM-based candidate filtering mechanism, substantially outperforms existing strong baselines in both design compliance and interactive functionality. Ultimately, PaperVoyager offers a new paradigm for scientific paper understanding, shifting the reading experience from passive comprehension to interactive exploration.

\section{Limitations}

Despite the promising results, our work has several limitations. First, constructing a high-quality benchmark requires substantial computational resources, annotation effort, and expert review, making the overall cost high. As a result, we did not further expand the benchmark to cover more paper content, nor did we build a dedicated training set to support model training or fine-tuning. Second, our experiments mainly focus on evaluating representative VLMs under a fixed evaluation setting. More extensive experiments, including broader model comparisons, ablation studies, and larger-scale evaluations, remain for future work.




\bibliography{custom}
\bibliographystyle{acl_natbib}

\clearpage
\appendix

\section{Implementation Details}

\paragraph{Mechanism Extraction. }
Before generating the structured specification, PaperVoyager first parses the input PDF to obtain a mechanism-aware understanding of the paper. Rather than treating the document as plain text, PaperVoyager uses a multimodal parsing module to convert PDF pages into a structured representation that preserves text, figures, equations, algorithmic pseudocode, and system diagrams. This step is essential for technical papers, where key mechanisms are often expressed through both textual explanations and visual elements. To reduce unnecessary context while retaining implementation-relevant information, the parser prioritizes sections such as methodology, algorithm descriptions, system design, and experiments, and de-emphasizes peripheral content such as related work. Based on the resulting multimodal representation, PaperVoyager identifies the paper’s core mechanisms, including step-by-step algorithmic procedures, parameter-dependent behaviors, state transitions, and relationships among system components. This mechanism-aware extraction bridges the gap between static PDF content and executable interactive system design.

\paragraph{Structured Specification Generation. }
Instead of directly generating code from the parsed paper, PaperVoyager first asks the model to determine which mechanisms should be exposed through interactive visualization. The model identifies key presentation points suitable for dynamic exploration, such as algorithm execution steps, parameter-controlled behaviors, and state changes, and converts them into a structured specification. This specification enumerates the required interactive modules, user controls, and expected visual outputs. The resulting specification includes a numbered \emph{Module Plan}, which decomposes the paper into a set of self-contained interactive modules before code generation begins. In this way, the Module Plan serves as a code-oriented intermediate representation that guides subsequent block-level generation.

\paragraph{Block-Level Generation. }
Conditioned on the specification, PaperVoyager decomposes the application into
independent \emph{blocks}, one per module in the Module Plan.  Each block is
generated as a standalone React/TypeScript component that implements only the
functionality of that module.  To promote diversity while controlling quality,
each block is generated $k$ times, producing $k$ candidate
implementations per block.  This block-level decomposition reduces per-call
complexity compared to monolithic whole-app generation, allowing the model to
focus on one interactive mechanism at a time.

\paragraph{Candidate Filtering.}
Each candidate component is compiled through a standard web build pipeline
(Vite + React) and rendered in a headless browser to produce a screenshot.  A
lightweight vision-language model (Qwen3-VL-4B-Instruct) then evaluates each
screenshot against its block specification by asking whether the rendered output
constitutes a functional and visually complete implementation of the
corresponding module.  Rather than using the model's discrete binary prediction,
we extract the raw logits of the \textit{Yes} and \textit{No} output tokens and
use their ratio as a continuous quality score, enabling fine-grained ranking of
all $k$ candidates.  The highest-scoring variant is retained for each block.

\paragraph{Block Merging.}
The selected best variants are unified into a final application by a
higher-capacity language model.  A merge prompt supplies all retained block
implementations together with the original specification and layout requirements.
The model synthesizes a single, self-contained \texttt{App.tsx} that integrates
every module under a shared sidebar navigation interface with a consistent visual
design.  The resulting file is compiled and served as the final interactive
website.

\begin{figure*}[!ht]
    \centering

    \begin{subfigure}[b]{0.32\textwidth}
        \centering
        \includegraphics[width=\textwidth]{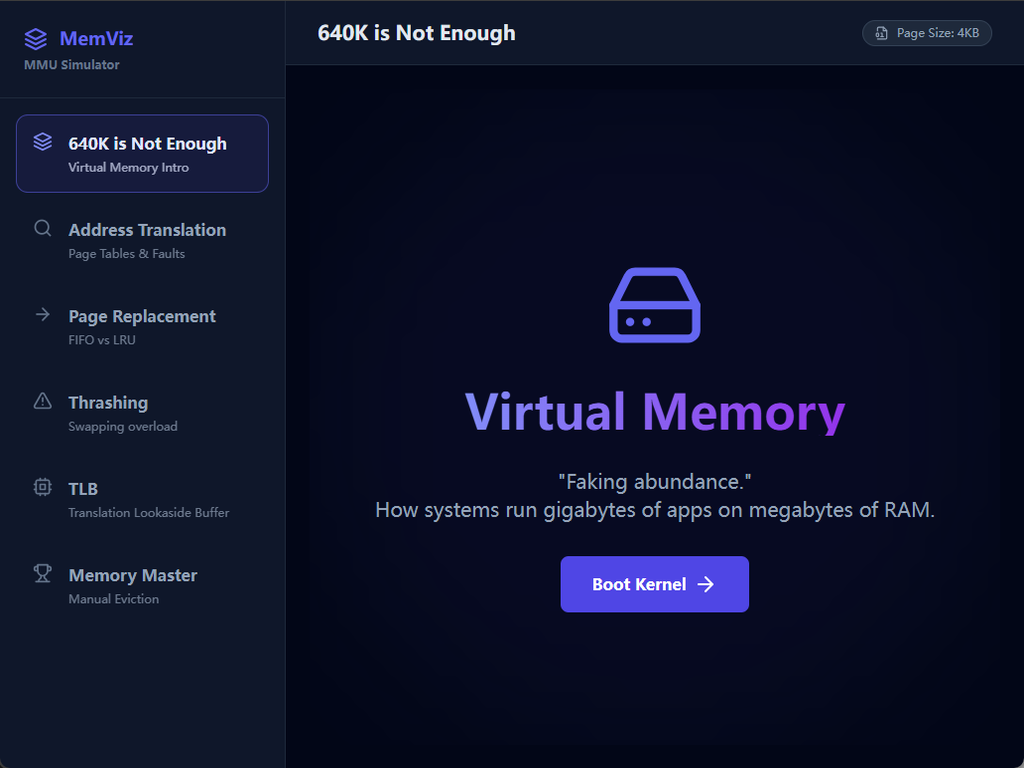}
        \label{fig:img2}
    \end{subfigure}
    \begin{subfigure}[b]{0.32\textwidth}
        \centering
        \includegraphics[width=\textwidth]{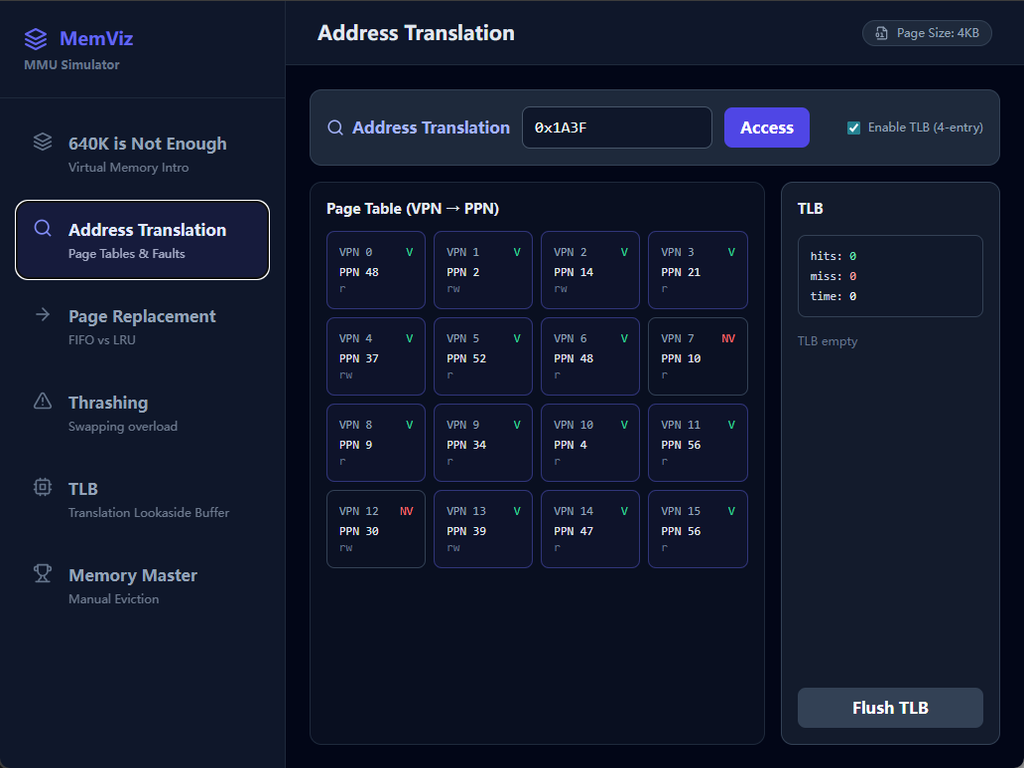}
        \label{fig:img1}
    \end{subfigure}
    \begin{subfigure}[b]{0.32\textwidth}
        \centering
        \includegraphics[width=\textwidth]{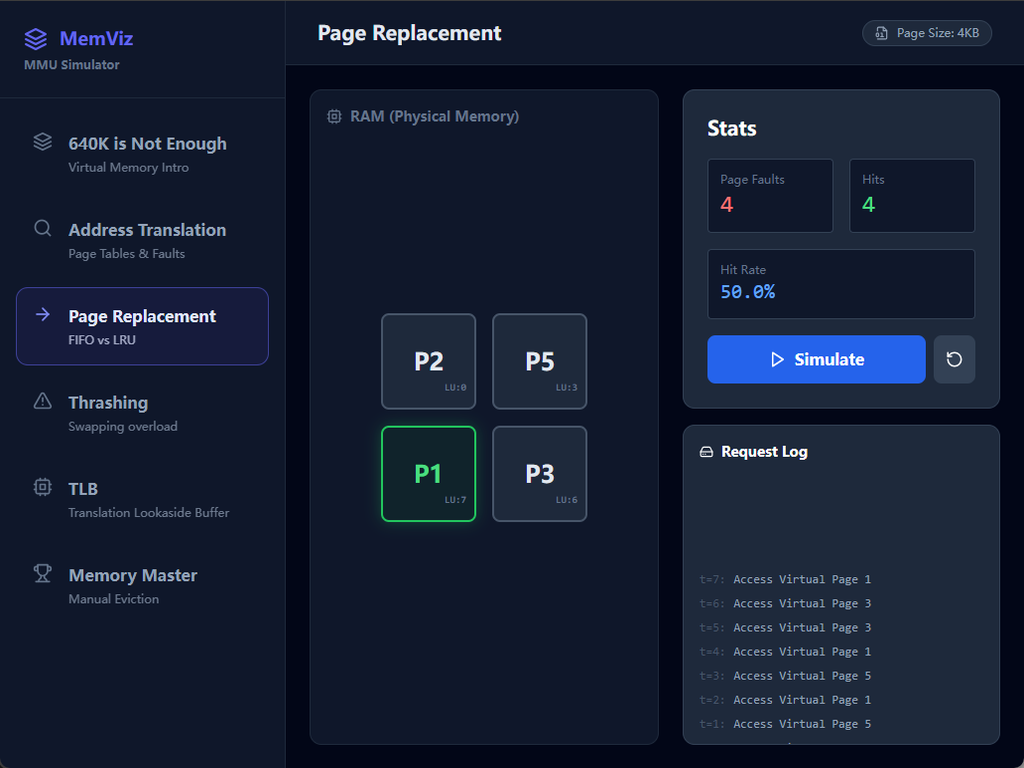}
        \label{fig:img3}
    \end{subfigure}

    \vspace{-4mm}
    \begin{subfigure}[b]{0.32\textwidth}
        \centering
        \includegraphics[width=\textwidth]{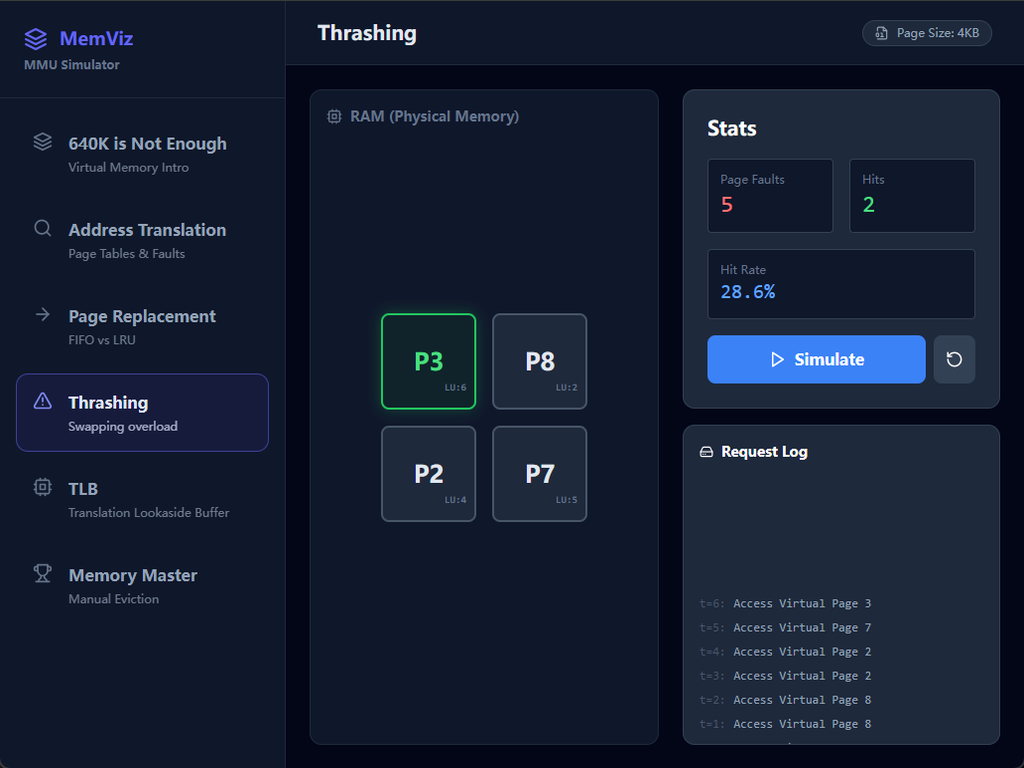}
        \label{fig:img3}
    \end{subfigure}
    \begin{subfigure}[b]{0.32\textwidth}
        \centering
        \includegraphics[width=\textwidth]{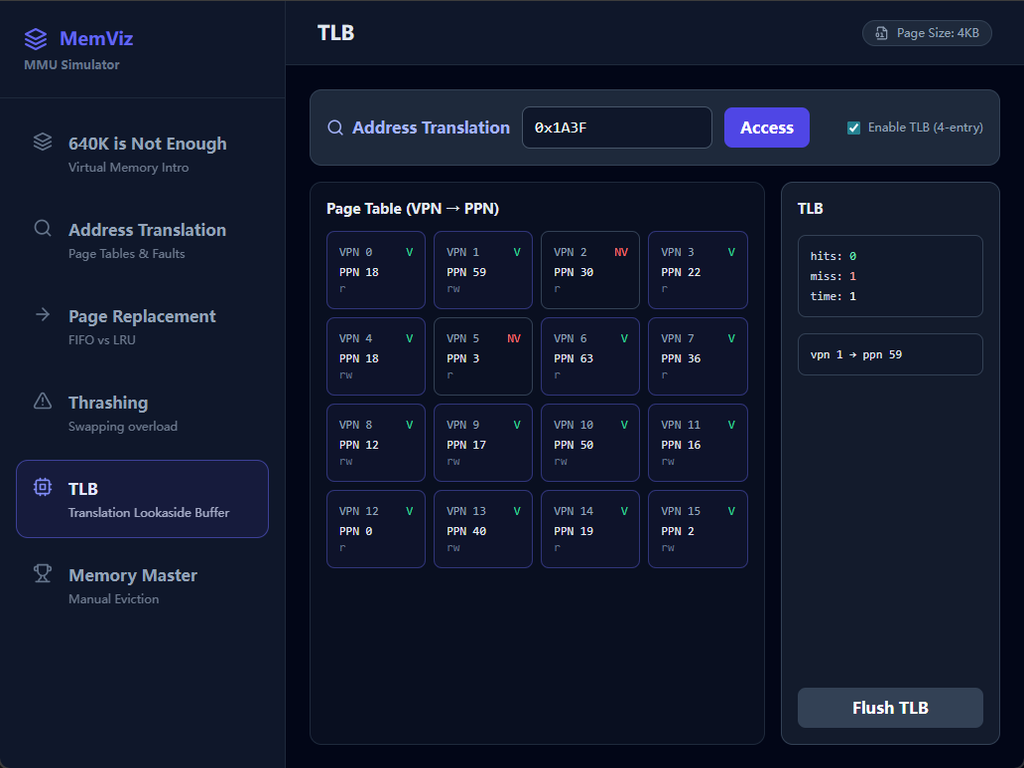}
        \label{fig:img2}
    \end{subfigure}
    \begin{subfigure}[b]{0.32\textwidth}
        \centering
        \includegraphics[width=\textwidth]{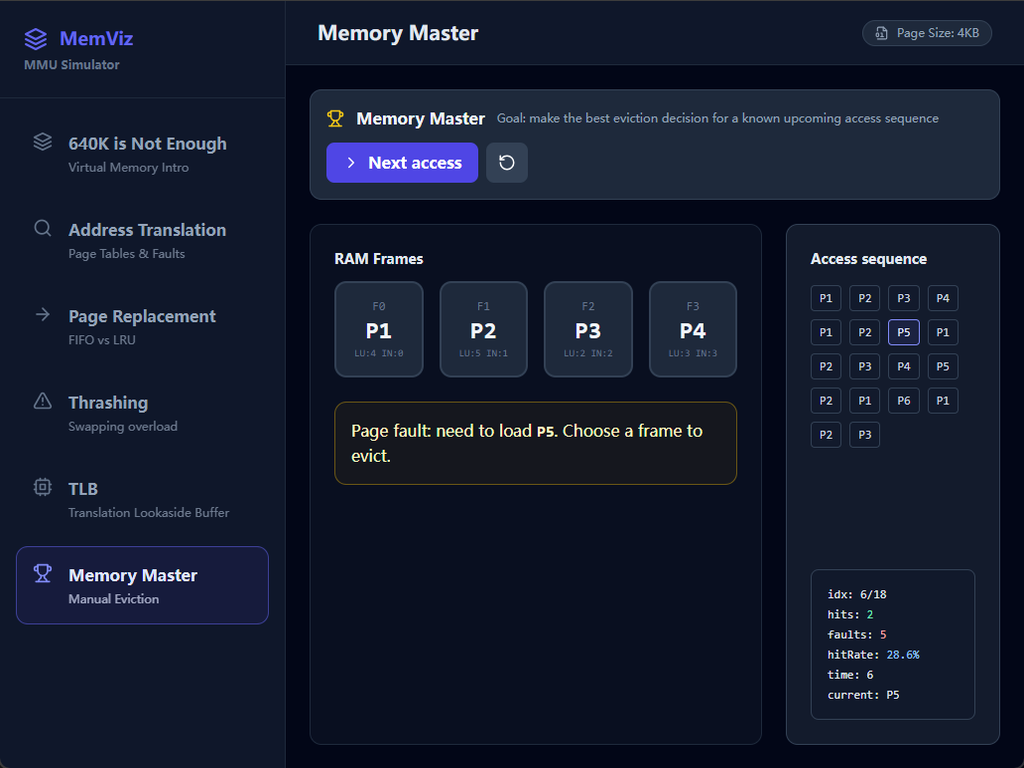}
        \label{fig:img1}
    \end{subfigure}
    
    \vspace{-5mm}
    \caption{A complete trajectory of a WebDemo example from our collected dataset. This example corresponds to a single-page interactive website on virtual memory, comprising multiple modules such as address translation, page replacement, thrashing, TLB simulation, and memory management. It demonstrates the target form of a completed WebDemo instance in our benchmark.}
    \label{fig:six_images1}
\end{figure*}

\begin{table*}[!ht]
\centering
\small
\setlength{\tabcolsep}{3pt}

\label{tab:benchmark_topics_all}
\begin{tabular}{@{}llll@{}}
\toprule
\textbf{Abbrev.} & \textbf{Topic} & \textbf{Domain} & \textbf{Originating Work} \\
\midrule
Alg-DP      & Dynamic Programming        & Algorithms        & \citet{bellman1957dynamic} \\
Alg-GP      & Graph Pathfinding          & Algorithms        & \citet{dijkstra1959note};
\citet{hart1968formal} \\
Alg-SR      & Sorting Algorithms         & Algorithms        & \citet{hoare1962quicksort};
\citet{williams1964heapsort} \\
DS-BT       & Balanced BSTs              & Data Structures   & \citet{avl1962}; \citet{guibas1978dichromatic}
\\
DS-HM       & Hash Maps / Cuckoo Hashing & Data Structures   & \citet{pagh2004cuckoo} \\
Dist-Raft   & Raft Consensus             & Distributed Sys.  & \citet{ongaro2014raft} \\
Math-Lorenz & Lorenz Attractor           & Mathematics       & \citet{lorenz1963deterministic} \\
Math-FFT    & Fourier Series / FFT       & Mathematics       & \citet{cooley1965fft} \\
Math-Eig    & Eigendecomposition         & Mathematics       & \citet{mises1929praktische} \\
Math-MC     & Monte Carlo Estimation     & Mathematics       & \citet{metropolis1949monte} \\
ML-GD       & Gradient Descent           & Machine Learning  & \citet{robbins1951stochastic};
\citet{kingma2015adam} \\
ML-KM       & K-Means Clustering         & Machine Learning  & \citet{macqueen1967kmeans};
\citet{arthur2007kmeanspp} \\
ML-NNV      & Neural Net Backprop        & Machine Learning  & \citet{rumelhart1986learning} \\
Phys-CFD    & 2D Fluid Simulation        & Physics           & \citet{stam1999stable} \\
Phys-Orbit  & N-body Gravity             & Physics           & \citet{verlet1967computer} \\
Phys-Opt    & Optics \& Ray Tracing      & Physics           & \citet{born1999principles} \\
Phys-Therm  & Thermodynamics             & Physics           & \citet{clausius1850ueber} \\
Sys-Sched   & CPU Scheduling             & Systems           & \citet{corbato1962experimental} \\
Sys-VM      & Virtual Memory \& Paging   & Systems           & \citet{denning1970virtual};
\citet{belady1966study} \\
\bottomrule
\end{tabular}
\caption{Benchmark topics, originating works, and domain categories. }
\end{table*}

\newpage

\section{Addressing Potential Evaluation Bias and Human Alignment}

To assess whether the original evaluation method exhibits systematic bias (e.g., unintentionally favouring a particular model) or deviates from human judgments, we conducted an additional controlled experiment on the Phys-Opt test case. We employed an independent strong evaluator (Claude) to re‑score all model outputs under identical conditions. Additionally, we collected human judgments from two expert annotators who were blind to model identity on the same test case. Table~\ref{tab:model_scores_physopt} reports the averaged scores from the original evaluation, Claude evaluation, and human evaluation on Phys-Opt

\begin{table}[!htbp]
\centering
\caption{Model scores under different evaluators}
\begin{tabular}{lccc}
\toprule
\textbf{Model} & \textbf{Original} & \textbf{Claude} & \textbf{Human} \\
\midrule
PaperVoyager & \textbf{85.0} & \textbf{73.1} & \textbf{79.2} \\
Gemini-3-Pro & \underline{81.3} & \underline{68.2} & \underline{75.0} \\
ChatGPT-5.2 & 71.9 & 57.5 & 64.5 \\
MiniMax & 59.5 & 42.8 & 51.0 \\
Qwen-2.5 & 55.3 & 46.2 & 48.3 \\
Kimi-K2 & 38.1 & 23.9 & 30.0 \\
\bottomrule
\label{tab:model_scores}
\end{tabular}
\end{table}

\textbf{Consistency across evaluators on Phys-Opt.} Despite absolute score differences (Claude tends to assign lower scores than humans), the relative ordering of models is highly consistent across all three evaluators. Spearman’s rank correlation between the original evaluation and human judgment is \( \rho = 0.99 \) (\(p<0.01\)), and between Claude and human judgment is \( \rho = 0.96 \) (\(p<0.01\)). This indicates that both automatic evaluation methods preserve the true performance ranking on this test case.

\textbf{No evidence of model‑specific bias or “score protection”.} If the original evaluation were biased toward any particular model, that model would show an anomalous gap between its original score and the human score compared to other models. However, the differences between original and human scores are of similar magnitude across all models, with no single model exhibiting an extreme deviation. Furthermore, the independent Claude evaluator produces scores that are uniformly lower than human scores, and the relative gaps between models remain consistent. Thus, on the Phys-Opt test case, there is no evidence that the original evaluation unfairly favours any specific model or suffers from misalignment with human judgments.

\textbf{Human alignment.} The Claude evaluation shows strong linear correlation with human judgments (Pearson’s \(r = 0.94\)), albeit with a negative bias (Claude is stricter). Most importantly, all three evaluation paradigms agree on the top‑performing model and the ranking of the bottom models on Phys-Opt. Therefore, our conclusions drawn from this test case are robust against the choice of evaluator, and there is no indication of systematic bias or poor alignment with human ratings.

\end{document}